\documentclass{article}
\usepackage{spconf,amsmath,graphicx}
\usepackage{multirow}
\usepackage{cite}
\usepackage{booktabs}
\usepackage{amssymb}
\usepackage{array}
\usepackage{xspace}
\usepackage{caption}
\usepackage{subcaption}

\usepackage{textpos}

\newcommand{\speechclip}{SpeechCLIP\xspace}
\newcommand{\hubert}{HuBERT\xspace}
\newcommand{\coco}{SpokenCOCO\xspace}
\newcommand{\flickr}{Flickr8k\xspace}
\newcommand{\fastvgs}{FaST-VGS\xspace}
\newcommand{\fastvgsp}{FaST-VGS+\xspace}

\title{SpeechCLIP: Integrating Speech\\with Pre-Trained Vision and Language Model}
\name{
    \begin{tabular}{c}
        Yi-Jen Shih$^{\text{1}}$, Hsuan-Fu Wang$^{\text{1}}$, Heng-Jui Chang$^{\text{1,2}}$, Layne Berry$^{\text{3}}$, Hung-yi Lee$^{\text{1}}$, David Harwath$^{\text{3}}$
    \end{tabular}
}
\address{
    $^{\text{1}}$National Taiwan University\\
    $^{\text{2}}$MIT CSAIL\\
    $^{\text{3}}$The University of Texas at Austin
}
\begin{document}
% \ninept
%
\maketitle
\begin{abstract}
Data-driven speech processing models usually perform well with a large amount of text supervision, but collecting transcribed speech data is costly.
Therefore, we propose \speechclip, a novel framework bridging speech and text through images to enhance speech models without transcriptions.
We leverage state-of-the-art pre-trained \hubert and CLIP, aligning them via paired images and spoken captions with minimal fine-tuning.
\speechclip outperforms prior state-of-the-art on image-speech retrieval and performs zero-shot speech-text retrieval without direct supervision from transcriptions.
Moreover, \speechclip can directly retrieve semantically related keywords from speech.
\end{abstract}
% for arxiv
\begin{textblock*}{\textwidth}(0cm,11cm)
\tiny
\noindent
Copyright 2022 IEEE. Published in the 2022 IEEE Spoken Language Technology Workshop (SLT) (SLT 2022), scheduled for 19-22 January 2023 in Doha, Qatar. Personal use of this material is permitted. However, permission to reprint/republish this material for advertising or promotional purposes or for creating new collective works for resale or redistribution to servers or lists, or to reuse any copyrighted component of this work in other works, must be obtained from the IEEE. Contact: Manager, Copyrights and Permissions / IEEE Service Center / 445 Hoes Lane / P.O. Box 1331 / Piscataway, NJ 08855-1331, USA. Telephone: + Intl. 908-562-3966.
\end{textblock*}
\section{Introduction}
\label{sec:intro}

Conventionally, speech processing tasks like speech recognition need transcribed speech data for machine learning.
They usually require large labeled datasets to perform well, but transcribing an enormous amount of speech is expensive.
Therefore, recent studies exploit unlabeled speech to pre-train models with self-supervised learning (SSL)~\cite{mohamed2022sslreview}.
Models learn to predict pseudo targets generated from raw data in SSL pre-training.
Some typical speech SSL methods include masked reconstruction~\cite{liu2020mockingjay,liu2021tera,chung2019apc,chung2020vq-apc,liu2021npc}, contrastive learing~\cite{oord2018cpc,schneider2019wav2vec,baevski2020vq-wav2vec,baevski2020wav2vec2,chung2021w2v}, classification~\cite{hsu2021hubert,chen2021wavlm,chiu2022bestrq}, multi-task learning~\cite{ravanelli2020paseplus}, and knowledge distillation~\cite{chang2022distilhubert,baevski2022data2vec,wang2022lighthubert}.
These methods succeed in a wide range of speech processing problems~\cite{yang2021superb,evain2021lebenchmark,tsai2022superbsg}.

Besides SSL methods focusing on a single modality, researchers propose using data from other modalities to boost machine performance on a specific modality.
E.g., pairing images with semantically related text or spoken captions is a typical method since collecting parallel image-text or image-speech data is fast and inexpensive~\cite{harwath2018jointly}.
Specifically, paired image-text data can be obtained by crawling images and captions from the internet.
Paired image-speech data can be collected by uttering text captions or describing images in speech.
This paper uses paired image-speech data and an image-text pre-trained model to enhance speech SSL models.

\begin{figure}
    \centering
    \includegraphics[width=0.75\linewidth]{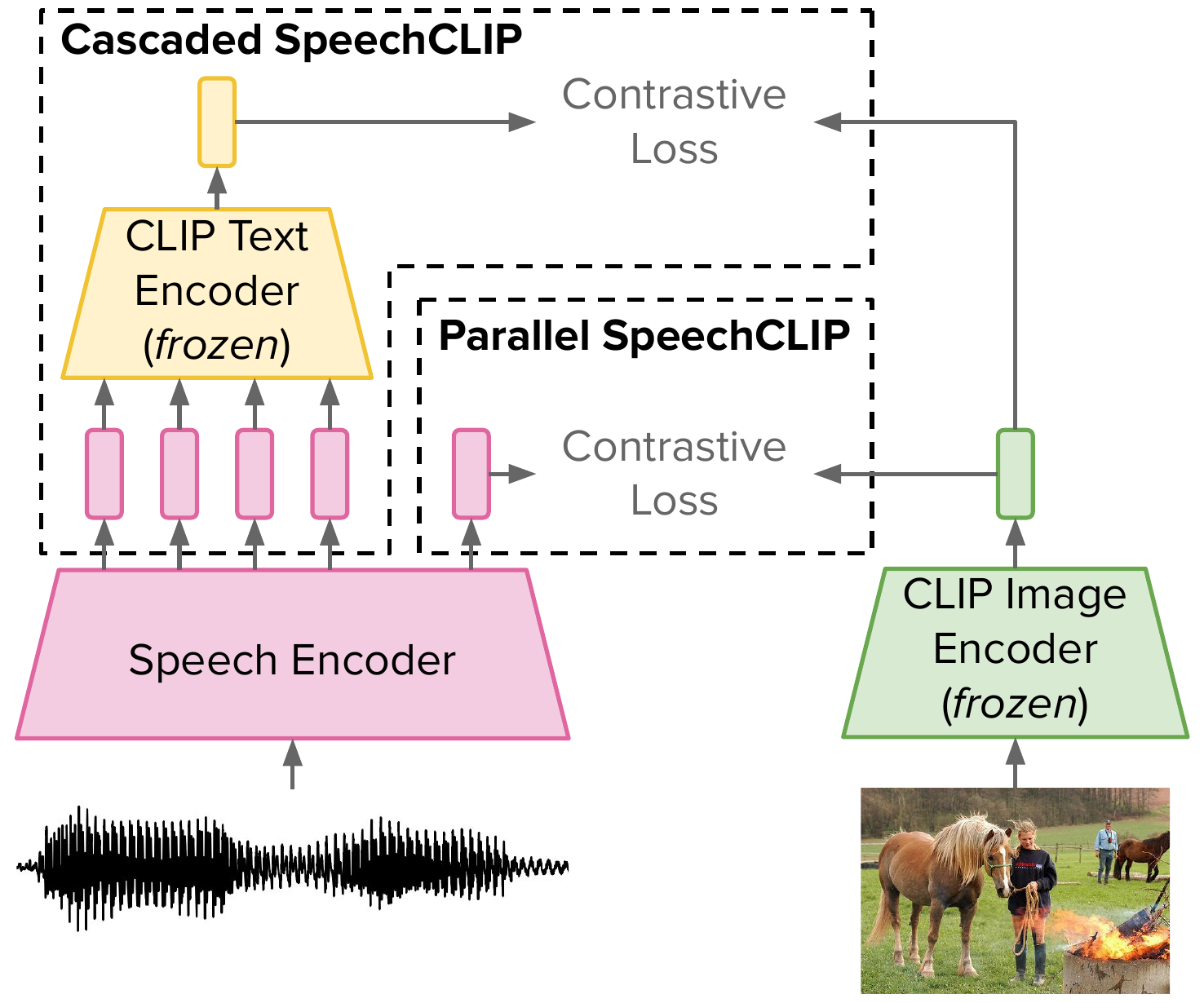}
    \vspace{-4pt}
    \caption{
        An overview of the proposed \speechclip model.
    }
    \label{fig:model_summ}
    \vspace{-10pt}
\end{figure}

\begin{figure*}
    \centering
    \begin{subfigure}[b]{0.48\textwidth}
		\centering
		\includegraphics[width=\textwidth]{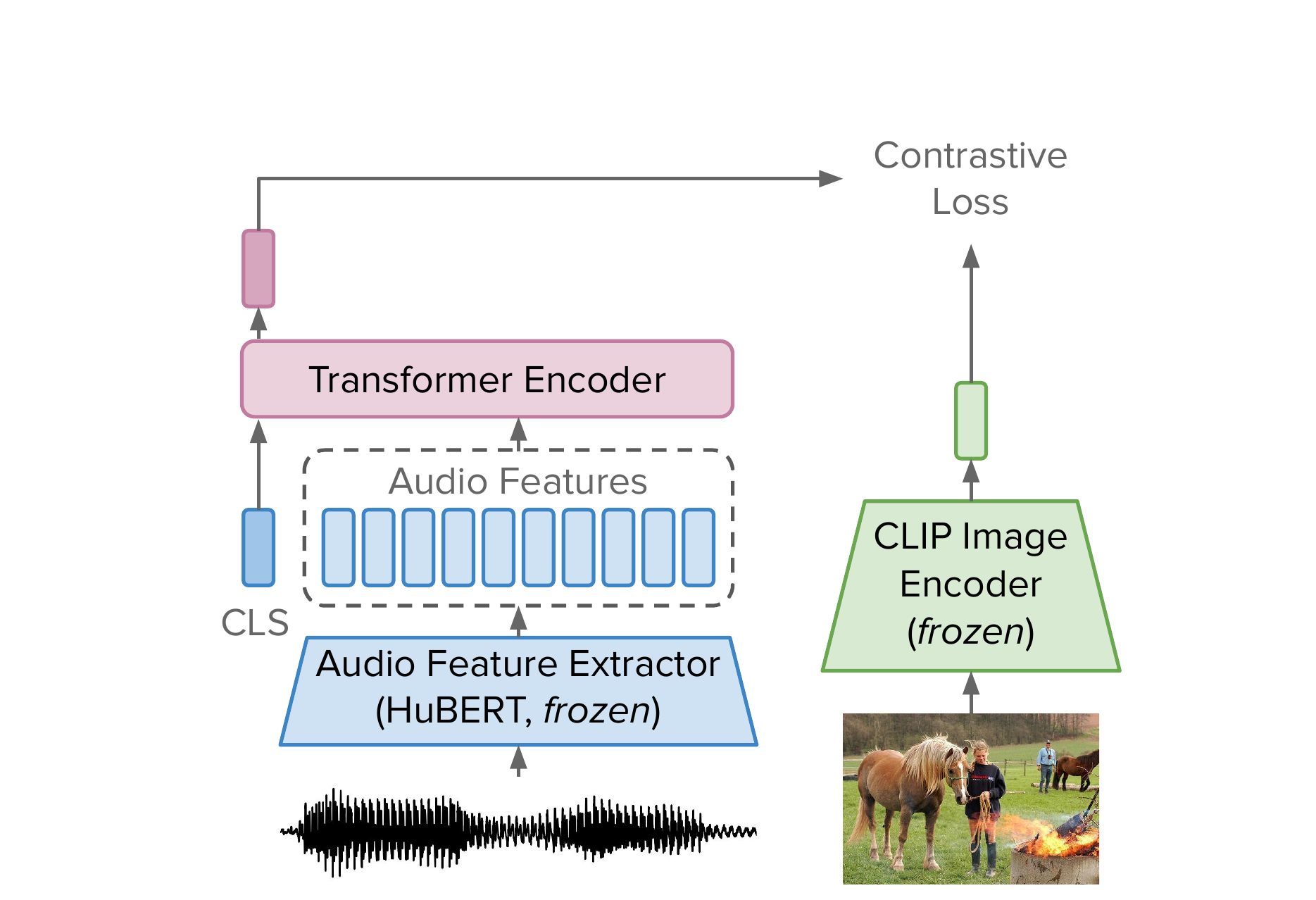}
		\caption{\normalsize Parallel \speechclip}
		\label{fig:model_para}
		\medskip
	\end{subfigure}
	\hfill
	\begin{subfigure}[b]{0.48\textwidth}
		\centering
		\includegraphics[width=\textwidth]{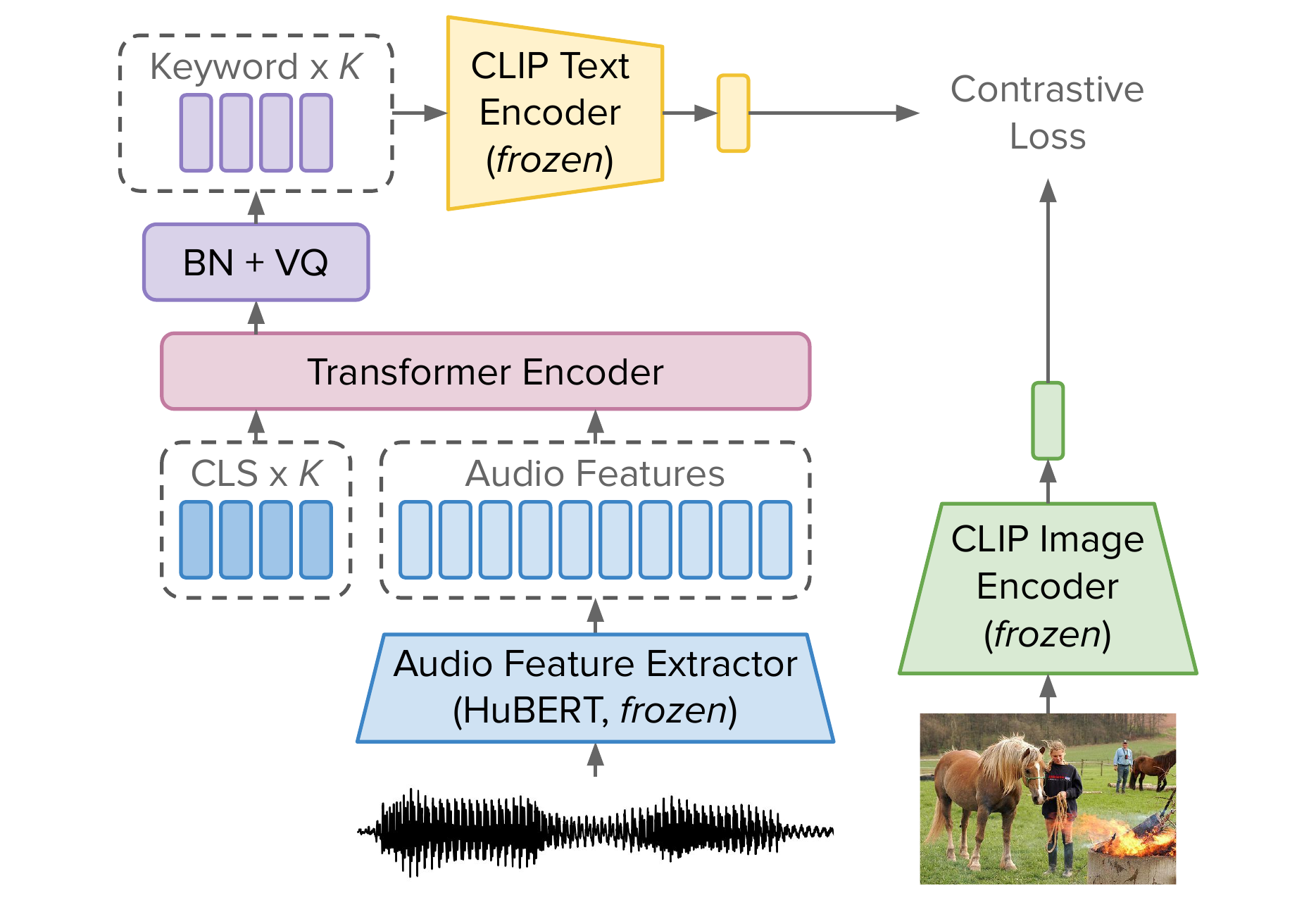}
		\caption{\normalsize Cascaded \speechclip}
		\label{fig:model_casc}
		\medskip
	\end{subfigure}
    \vspace{-10pt}
    \caption{
        An illustration of \speechclip models.
        (a) A pre-trained \hubert~\cite{hsu2021hubert} extracts audio features.
        The features are concatenated with a learnable CLS token and fed into a transformer encoder layer to obtain a single vector representing the information of the entire sequence.
        The vector is then used to compute contrastive loss with the CLIP image encoder's output~\cite{radford2021clip}.
        (b) Cascaded \speechclip uses $K$ CLS tokens to capture a small sequence of keywords from the audio signal.
        The keywords are batch-normalized and vector-quantized before passing to the CLIP text encoder.
        BN and VQ respectively denote batch normalization and vector quantization.
    }
    \label{fig:model_detail}
    \vspace{-8pt}
\end{figure*}

Much effort was put into using paired images and spoken captions to help speech processing~\cite{chrupala2022visually}, and they are usually called visually grounded speech models~(VGS).
VGS models benefit many applications like speech recognition~\cite{hsu2019transfer}, word discovery~\cite{harwath2015deep}, speech generation~\cite{hsu2020text}, cross-modal alignment~\cite{harwath2018jointly,wang2021align,khorrami2021evaluation}, and multilingual spoken language processing~\cite{harwath2018vision,kamper2018visually,havard2020catplayinginthesnow,ohishi2020trilingual}.
Most studies pre-train and evaluate VGS models on image-speech retrieval, showing the capabilities of capturing the correspondence between images and speech~\cite{ilharco2019large,sanabria2021milan}.
E.g., the recent Fast-Slow Transformer for Visually Grounding Speech~(\fastvgs and \fastvgsp) succeeds in many speech processing tasks by utilizing transformers and cross-modal attention mechanisms to perform image-speech retrieval and semantic tasks~\cite{peng2022fastvgs,peng2022fastvgsplus}.
Moreover, VGS models trained with retrieval objectives can extract semantic and word-level information from speech~\cite{peng2022vghubert}, which is difficult to achieve by training solely with speech~\cite{pasad2021layer}.

While many studies obtain semantic information from speech without transcriptions, some extent of assistance from text could be helpful for some tasks.
E.g., recent unsupervised ASR methods rely on nonparallel text data and a pronunciation lexicon~\cite{baevski2021wav2vec-u,liu2022wav2vec-u2}.
To circumvent transcriptions or lexicons, we propose to bridge speech and text domains via images, i.e., taking advantage of paired image-speech and image-text data.
Thus, this paper introduces \speechclip, a novel framework to integrate speech SSL models with a pre-trained vision and language model as depicted in Fig.~\ref{fig:model_summ}.
We use Contrastive Language-Image Pre-training~(CLIP), a powerful model pre-trained to align parallel image-text data~\cite{radford2021clip}.
Then, a speech encoder initialized by a pre-trained speech SSL model is enhanced by aligning with CLIP using paired image-speech data.
By aligning a speech encoder's and CLIP's image embedding spaces, the speech encoder is implicitly aligned with CLIP's text encoder, forcing it to capture more textual content.

We propose two \speechclip architectures: parallel and cascaded.
The parallel model is similar to WAV2CLIP~\cite{wu2022wav2clip}.
However, our speech encoder uses a pre-trained speech SSL model and focuses on capturing local and global spoken contents.
Meanwhile, WAV2CLIP extracts global features in general audio for classification and retrieval.
Furthermore, AudioCLIP is an extension of WAV2CLIP since it is trained with paired image, audio, and text data~\cite{guzhov2022audioclip}.
The cascaded \speechclip cascades CLIP's text encoder on top of the speech encoder, forcing the model to output subword embeddings.
Eventually, the cascaded model captures spoken words in speech signals.

In this paper, the proposed \speechclip models achieve state-of-the-art image-speech retrieval on two standard spoken caption datasets with minimal fine-tuning.
Moreover, we demonstrate \speechclip's capability of performing zero-shot speech-text retrieval and capturing keywords directly from speech.
We also make our code available on Github\footnote{https://github.com/atosystem/SpeechCLIP}.

\section{Method}
\label{sec:method}

\subsection{Preliminaries}
\label{subsec:prelim}

We briefly explain pre-trained models used in \speechclip.

\noindent\textbf{Contrastive Language-Image Pre-training (CLIP)~\cite{radford2021clip}.}
CLIP uses contrastive learning to pre-train visual models from natural language supervision on an enormous scale, where the supervision comes from paired image-text data.
Composing two encoders processing image and text separately, CLIP aims to align semantically similar images and text captions.
CLIP can easily transfer across various computer vision tasks with little supervision.

\noindent\textbf{Hidden-unit BERT (\hubert)~\cite{hsu2021hubert}.}
\hubert is a speech SSL method similar to masked language modeling, predicting labels generated by clustered acoustic features.
\hubert comprises a CNN feature extractor followed by a transformer encoder~\cite{vaswani2017attention} and offers good initialization for many speech processing tasks~\cite{yang2021superb,tsai2022superbsg}.

In \speechclip, pre-trained CLIP and \hubert models are frozen and serve as feature extractors, as shown in Fig.~\ref{fig:model_detail}.
The CLIP model extracts image and sentence embeddings to supervise \speechclip.
Following SUPERB~\cite{yang2021superb}, \hubert's CNN output and transformer encoder's hidden representations are weighted and summed by a set of learnable weights.
The weights automatically assign importance to each hidden layer to minimize the overall objective function.
Only the newly added components excluding \hubert and CLIP are learnable during training, reducing the computational cost significantly, thus enabling a larger batch size for contrastive pre-training.
In the following sections, we introduce two \speechclip architectures: parallel and cascaded.

\subsection{Parallel \speechclip}
\label{subsec:parallel}

Parallel \speechclip is similar to CLIP, which aligns semantically related images and spoken captions, as shown in Fig.~\ref{fig:model_para}.
Since the weighted sum of \hubert's output is a sequence of frame-level features, we add a learnable CLS token at the beginning of each sequence.
The sequence is passed through a transformer encoder layer to obtain an utterance-level representation~\cite{vaswani2017attention}.
The representation is used to compute the cosine similarity with image embeddings in a mini-batch for calculating the contrastive loss.
Cosine similarity scores are also used for retrieving speech and image samples.
Following CLIP, the loss function has a learnable temperature for scaling the similarity scores.

By aligning speech and CLIP image encoders, parallel \speechclip implicitly bridges speech and text representations since CLIP's image and text encoders are well-aligned.
Therefore, it can perform both image-speech and speech-text retrieval.
Still, this method is limited to summarizing utterances because it has no explicit constraints to capture word-level content.
Thus, the following section introduces a novel method addressing this issue.

\subsection{Cascaded \speechclip}
\label{subsec:cascaded}

To force the speech encoder to capture semantic information from speech, we propose cascaded \speechclip by cascading speech encoder with CLIP's text encoder as shown in Fig.~\ref{fig:model_casc}.
Following parallel \speechclip, the cascaded model is trained with contrastive loss, but the difference lies in the summarization process of utterances.

First, we add $K$ learnable CLS tokens at the beginning of an audio feature sequence, where $K$ is a hyper-parameter for the number of keywords obtained from an utterance.
The sequence is fed into a transformer encoder and projected to the CLIP input embedding dimension.
Next, the projected CLS tokens are batch-normalized to match the mean and variance of CLIP's subword embeddings.
We apply vector quantization (VQ) to map the $K$ normalized embeddings to CLIP's $V$ subword embeddings.
This operation produces keywords indicating the essential concepts in each utterance.

The VQ process is described as follows.
We first compute the cosine similarity between the $k^{\text{th}}$ normalized CLS embedding ($\boldsymbol{z}_k$) and the $v^{\text{th}}$ subword embedding ($\boldsymbol{e}_v$) as
\vspace{-3pt}
\begin{equation}
    \vspace{-3pt}
    s_{kv} = \text{cos}\left( \boldsymbol{z}_k , \boldsymbol{e}_v \right).
    \label{eq:cos}
\end{equation}
Next, we choose the subword embedding with the highest similarity from the vocabulary, which can be expressed as
\vspace{-3pt}
\begin{equation}
    \vspace{-3pt}
    \boldsymbol{e}_{v^{\star}}, \text{~where~} v^{\star} = \underset{1 \leq v \leq V}{\text{argmax}} ~ s_{kv}.
    \label{eq:hardkw}
\end{equation}
Since $\boldsymbol{e}_{v^{\star}}$ is not differentiable, we compute another embedding by weighted summing all $V$ subword embeddings as
\vspace{-3pt}
\begin{equation}
    \vspace{-3pt}
    \overline{\boldsymbol{h}}_k = \left[ \boldsymbol{e}_1 \dots \boldsymbol{e}_V \right] \text{softmax}\left( \left[ s_{k1} \dots s_{kV} \right]^\top / \tau \right),
    \label{eq:softkw}
\end{equation}
where each embedding $\boldsymbol{e}_v$ is a column vector and $\tau$ is a hyper-parameter ($\tau=0.1$).
Combining Eqs. \ref{eq:hardkw} and \ref{eq:softkw}, we apply straight-through gradient estimator~\cite{bengio2013estimating} to obtain quantized keywords
\vspace{-3pt}
\begin{equation}
    \vspace{-3pt}
    \boldsymbol{h}_k = \boldsymbol{e}_{v^{\star}} + \overline{\boldsymbol{h}}_k - \text{sg}\left( \overline{\boldsymbol{h}}_k \right),
\end{equation}
where $\text{sg}(x) = x$ and $\frac{d}{dx}\text{sg}(x)=0$ is the stop gradient operator.
The $K$ keywords are then fed into the CLIP text encoder for computing the contrastive objective.

Overall, the cascaded \speechclip encourages the speech encoder to extract subwords because of the supervision from the CLIP text encoder.
Hence, it is expected to capture more semantic and content information from speech.

% \subsection{Hybrid \speechclip}
% \label{subsec:hybrid}

% Besides the two architectures introduced above, we propose combining them as a hybrid model because they serve different purposes.
% Parallel \speechclip aims to summarize an utterance to a single vector, while cascaded \speechclip learns to capture multiple spoken keywords in an utterance.
% Therefore, we expect combining the two architectures to improve performance on downstream tasks.

% \ian{
% In the hybrid \speechclip model, parallel and cascaded branches have the same structure as parallel and cascaded settings respectively, but share the same \hubert model.
% % MHSA layer 
% % but use separate CLS tokens, composing $K + 1$ CLS tokens.
% The objective function is the sum of the contrastive losses computed separately from the two branches.}
\vspace{-5pt}
\section{Experiment}
\label{sec:experiment}

\begin{table}[t]
    \caption{
        Model details.
        The number of parameters varies since they include parallel and cascaded models.
    }
    % \vspace{2pt}
    \label{tab:model_compare}
    \centering
    \small
    % \footnotesize
    \begin{tabular}{@{~~}l@{~~}c@{~~}c@{~~}c@{~~}c@{~~}c@{~~}}
        \toprule
        & Audio & CLIP Image & Trainable & Total \\
        Model & Encoder & Encoder & Params (M) & Params (M) \\
        \midrule
        \multirow{2}{*}{Base}  & \hubert Base & ViT-B/32 & \multirow{2}{*}{2.8 -- 7.5} & \multirow{2}{*}{252 -- 257} \\
         & \scriptsize{(95 M)}   & \scriptsize{(250 M)}   \\
        \multirow{2}{*}{Large} & \hubert Large & ViT-L/14  & \multirow{2}{*}{6.1 -- 13.4}  & \multirow{2}{*}{765 -- 772} \\
        & \scriptsize{(316 M)}   &  \scriptsize{(422 M)} \\
        \bottomrule
    \end{tabular}
    \vspace{-8pt}
\end{table}

\subsection{Setup}
\label{subsec:setup}

\noindent\textbf{Dataset.}
\speechclip is pre-trained and evaluated with retrieval on \flickr Audio Captions Corpus~\cite{harwath2015deep} and \coco dataset~\cite{hsu2020text}.
Each image in both datasets is paired with five spoken captions produced by humans uttering text captions.
\flickr consists of 8k images and 46 hours of speech, while \coco has 123k images and 742 hours of speech.
Following \fastvgs, we use the Karpathy split for \coco~\cite{karpathy2015deep}.

\noindent\textbf{Model.}
We implemented \speechclip in two sizes: Base and Large, a detailed comparison is shown in Table \ref{tab:model_compare}.
Note that we omit the Base notation in the following sections.
The hidden dimension of the transformer encoder is the same as that of the audio encoder.
The feed-forward network in the cascaded model's transformer encoder is removed for better performance.
Parallel and cascaded models have respectively eight and one attention head.
We set $K$ to 8 in all experiments.
All models are trained with Adam optimizer with a weight decay of 10$^{-\text{6}}$, batch size of 256, and 50k steps in total.
The learning rate linearly increases to 10$^{-\text{4}}$ in the first 5k steps and linearly decreases to 10$^{-\text{8}}$ afterward.
All experiments are conducted on a 32GB V100 GPU except for pre-training on \coco, which uses two.
The largest model's pre-training lasts approximately two days.

\begin{table}[t]
    \centering
    \caption{
        Recall scores for image-speech retrieval on \flickr and \coco testing sets.
        % The subscripts after the hybrid models indicate which branch is used for calculating retrieval, where P and C respectively stands for parallel and cascaded.
    }
    \label{tab:recall}
    % \vspace{2pt}
    \small
    \begin{tabular}{@{~~}l@{~~}c@{~~}c@{~~}c@{~}c@{~}c@{~~}c@{~~}c@{~~}} 
        \toprule
        & \multicolumn{3}{c}{Speech $\rightarrow$ Image} & & \multicolumn{3}{c}{Image $\rightarrow$ Speech} \\ 
        \cmidrule{2-4}
        \cmidrule{6-8}
        Method & R@1 & R@5 & R@10 & & R@1 & R@5 & R@10 \\ 
        \midrule
        & \multicolumn{7}{c}{\flickr} \\
        \cmidrule{2-8}
        
        FaST-VGS$_{\text{CO}}$~\cite{peng2022fastvgs} & 26.6 & 56.4 & 68.8 & & 36.2 & 66.1 & 76.5 \\
        FaST-VGS$_{\text{CTF}}$~\cite{peng2022fastvgs} & 29.3 & 58.6 & 71.0 & & 37.9 & 68.5 & 79.9 \\
        MILAN~\cite{sanabria2021milan} &
        33.2 & 62.7 & 73.9 & & 49.6 & 79.2 & 87.5\\
        \midrule
        
        % small model
        % Parallel (MHA) & 27.8 & 57.7 & 70.6 & & 39.5 & 73.8 & 84.1 \\
        Parallel & 26.7 & 57.1 & 70.0 & & 41.3 & 73.9 & 84.2 \\
        Cascaded & 8.2  & 25.7 & 37.2 & & 14.1 & 34.5 & 49.2 \\

        % \midrule
        Parallel Large & \textbf{39.1} & \textbf{72.0} & \textbf{83.0} & & \textbf{54.5} & \textbf{84.5} & \textbf{93.2} \\
        % Parallel Large (MHA) & 40.5 &	73.4 & 84.7 & & 54.0 & 85.6 & 93.8 \\
        Cascaded Large & 14.7 & 41.2 & 55.1 & & 21.8 & 52.0 & 67.7 \\
 
        % \midrule
        % \midrule
        % CLIP ViT-B/32 & 54.7 & 80.5 & 88.5 & & 70.2 & 91.9 & 96.1 \\
        % CLIP ViT-L/14 & 61.3 & 85.9 & 92.6 & & 78.3 & 93.8 & 97.5 \\
        \midrule
        & \multicolumn{7}{c}{\coco} \\
        \cmidrule{2-8}
        ResDAVEnet~\cite{hsu2019transfer} &
        17.3 & 41.9 & 55.0 & & 22.0 & 50.6 & 65.2 \\
        FaST-VGS$_{\text{CO}}$~\cite{peng2022fastvgs} & 31.8 &  62.5 &  75.0 & & 42.5 &  73.7 & 84.9 \\
        FaST-VGS$_{\text{CTF}}$~\cite{peng2022fastvgs} & \textbf{35.9} &  66.3 &  77.9 & & 48.8 &  78.2 & 87.0 \\
        \midrule
        
        Parallel Large & 35.8 & \textbf{66.5} & \textbf{78.0} & & \textbf{50.6} & \textbf{80.9} & \textbf{89.1} \\
        % Parallel Large (MHA) & 35.6 &	64.3 &	75.9 & & 47.9 &	77.6 &	86.9 \\
        Cascaded Large & 6.4 & 20.7 & 31.0 & & 9.6 & 27.7 & 39.7 \\

        % \midrule
        % \hl{CLIP ViT-L/14} & 36.1  & 60.8  & 71.1 & & 56.1 & 78.8 & 86.9 \\
        \bottomrule
    \end{tabular}
    \vspace*{-10pt}
\end{table}

\subsection{Image-Speech Retrieval}
\label{subsec:retrieval}

In this section, we evaluate \speechclip on the image-speech retrieval task, showing how well models can align speech with CLIP image embeddings.
As shown in Table~\ref{tab:recall}, parallel \speechclip models surpass almost all baseline methods, especially for the Large \speechclip models.
The parallel Base model on \flickr also shows competitive performance with the \fastvgs$_{\text{CO}}$ model, indicating that utilizing powerful pre-trained models and a small set of learnable parameters is sufficient.
Moreover, the cascaded models obtain the lowest recall scores because passing encoded speech through VQ and a CLIP text encoder loses information.
Overall, the results show the benefits of integrating CLIP in VGS models even with minimal fine-tuning.

\begin{table}[t]
    \centering
    \caption{
        Recall for speech-text retrieval on \flickr and \coco.
        `Sup.'' indicates the supervised version of parallel \speechclip by replacing the image encoder with CLIP text encoder in parallel \speechclip.
    }
    \label{tab:zeroshot_speech_text}
    % \vspace{2pt}
    \small
    \begin{tabular}{@{~~}l@{~~}c@{~~}c@{~~}c@{~}c@{~}c@{~~}c@{~~}c@{~~}}
        \toprule
        & \multicolumn{3}{c}{Speech $\rightarrow$ Text}  & & \multicolumn{3}{c}{Text $\rightarrow$ Speech} \\ 
        \cmidrule{2-4}
        \cmidrule{6-8}
        Method & R@1 & R@5 & R@10 & & R@1 & R@5 & R@10 \\ 
        \midrule
        & \multicolumn{7}{c}{\flickr} \\
        \cmidrule{2-8}
        Random & 0.10 &	0.50 &	0.99 & & 0.10 &	0.50 &	0.99 \\
        Parallel Large & 19.56 & 44.06 & 58.46 & & 22.50 & 44.14 & 54.54 \\
        % Hybrid Large & & 64.58 & 85.03 &	90.73 &	& 64.62 &	85.24 &	90.52 \\
        % Hybrid Large & 63.40 & 84.34 & 90.16 & & 59.41 &	81.91 &	88.57 \\
        Parallel Large (Sup.) & 97.06 & 99.24 & 99.46 & & 97.88 & 99.76 & 99.90 \\
        \midrule
        & \multicolumn{7}{c}{\coco} \\
        \cmidrule{2-8}
        Random & 0.02 &	0.10 &	0.20 & & 0.02 &	0.10 &	0.20 \\
        Parallel Large & 60.32 & 81.81 & 88.18 & & 65.45 &	85.82 &	91.27 \\
        % Hybrid Large & & 64.58 & 85.03 &	90.73 &	& 64.62 &	85.24 &	90.52 \\
        % Hybrid Large & 63.40 & 84.34 & 90.16 & & 59.41 &	81.91 &	88.57 \\
        Parallel Large (Sup.) & 
        95.02 &	99.46 &	99.78 & & 95.35 &	99.68 &	99.93 \\
        \bottomrule
    \end{tabular}
\end{table}

\subsection{Zero-shot Speech-Text Retrieval}
\label{subsec:speech_text_ret}

This section highlights parallel \speechclip's capability to perform zero-shot speech-text retrieval.
Speech and text representations are respectively computed from a pre-trained parallel \speechclip's speech encoder and a CLIP text encoder.
The representations are then used to calculate cosine similarity scores for retrieval.
Although this problem has been studied for a while, prior studies require either paired speech-text training data~\cite{duquenne2021multimodal,khurana2022samu} or pretrained image tagger~\cite{kamper2018semantic}.

Additionally, two supervised parallel \speechclip models respectively trained with paired spoken and text captions in \flickr and \coco are considered as toplines.
These models' CLIP image encoders are replaced with CLIP text encoders to align speech and text explicitly.
When computing recall, we regard retrieving speech and text captions related to the same image as successful.
Therefore, results only show whether models retrieve semantically related samples, not exact matching of speech and transcriptions.

According to Table~\ref{tab:zeroshot_speech_text}, proposed \speechclip models yield considerably better performance than random retrieval, showing that speech and text embedding spaces are well aligned.
Specifically, parallel \speechclip performs better on this task when trained on a larger dataset like \coco.
Although the performance gap between the proposed methods and the supervised toplines remains, we show that bridging speech and text with image is possible and promising.

We demonstrate that parallel \speechclip retrieves noisy transcriptions for speech signals.
These transcriptions can then be used for supervised or semi-supervised speech recognition model training.
Furthermore, by replacing CLIP with Multilingual-CLIP\footnote{https://github.com/FreddeFrallan/Multilingual-CLIP}, we can retrieve noisy transcriptions of different languages, thus performing speech translation.

\subsection{Keyword Retrieval with Cascaded SpeechCLIP}
\label{subsec:kw_ret}

% \begin{figure*}
%     \centering
%     \includegraphics[width=\linewidth,trim={4cm 0.6cm 17cm 0.4cm},clip]{figure/kw_ret.pdf}
%     \vspace{-10pt}
%     \caption{
%         Three examples come from the SpokenCOCO test set.
%         For each of the example we show the attention weights for each keyword, and we also show the retrieved subwords on the left (sorted in decreasing cosine similarity).
%         Subwords are in bold type if they exist in the golden caption.
%         \ian{(I will put the retrieval results under each attention map to make the font size bigger}
%     }
%     \label{fig:kw_ret}
% \end{figure*}

\begin{figure*}
    \includegraphics[width=\linewidth,trim={7.5cm 0.6cm 20.9cm 0.4cm},clip]{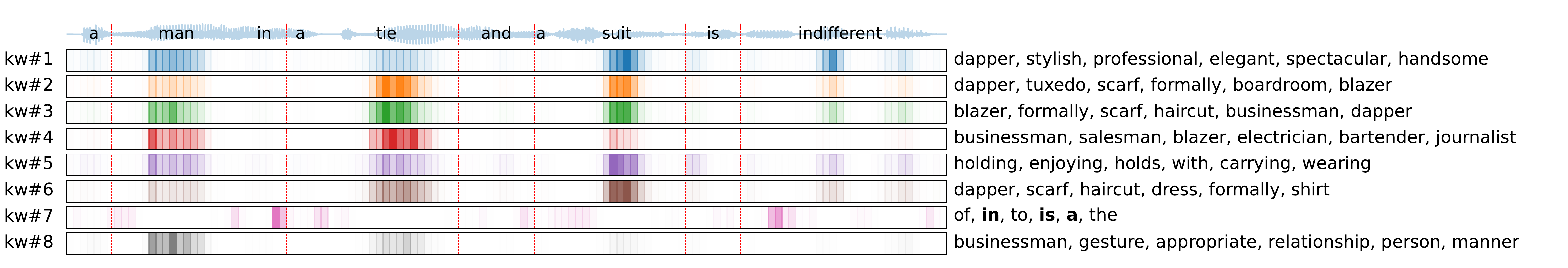}
    \\[\smallskipamount]
\includegraphics[width=\linewidth,trim={7.5cm 0.6cm 20.5cm 0.4cm},clip]{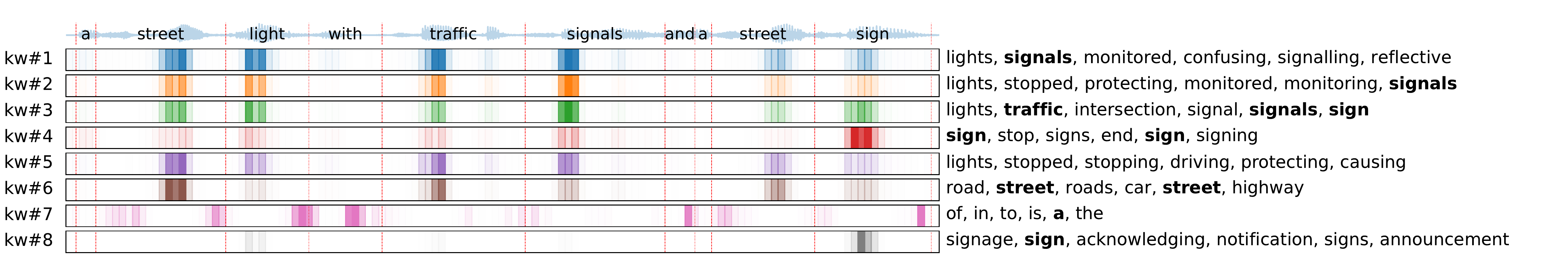}
    % \\[\smallskipamount]
    % \includegraphics[width=\linewidth,trim={7.5cm 0.6cm 20.5cm 0.4cm},clip]{figure/attention_maps/example3.pdf}
    \vspace{-20pt}
    \caption{
        Demonstration of a cascaded \speechclip Large model retrieving words using its CLS tokens' outputs.
        The two utterances are from the SpokenCOCO test set.
        For each keyword in each sample, we show the transformer encoder's attention map over the whole sequence and the retrieved subwords on the right and sorted in decreasing cosine similarity.
        Subwords in boldface indicate they exist in the ground truth caption.
    }
    \label{fig:kw_ret}
    \vspace{-8pt}
\end{figure*}

\begin{table}[t]
    \caption{
        Keyword hit rates for cascaded \speechclip.
        Avg denotes averaged hit rate.
        $^\dagger$ and $^\ddagger$ respectively denote models trained on \flickr and \coco.
    }
    % \vspace{2pt}
    \label{tab:kw_hit}
    \centering
    \small
    % \footnotesize
    \begin{tabular}{@{~~}l@{~~}c@{~~}c@{~~}c@{~~}c@{~~}c@{~~}c@{~~}c@{~~}c@{~~}c@{~~}}
        \toprule
        Model & kw1 & kw2 & kw3 & kw4 & kw5 & kw6 & kw7 & kw8 & Avg \\
        \midrule
        Base$^{\dagger}$ & 57.0 & 25.6 & 20.2 & 5.0 & 20.0 & 26.5 & 10.5 & 16.6 & 22.7 \\
        Large$^{\dagger}$ & 56.5 & 19.6 & 20.5 & 37.5 & 21.7 & 34.6 & 26.4 & 44.7 & 32.7\\
        Large$^{\ddagger}$ & 27.5 & 22.4 & 35.8 & 61.0 & 21.6 & 54.2 & 60.1 & 22.9 & 38.2 \\
        
        % Base$_{\dagger}$ & 57.0 & 26.5 & 25.6 & 20.2 & 20.0 & 16.6 & 10.5 & 5.0 & 22.7 \\
        % Large$^{\dagger}$ & 56.5 & 44.7 & 37.5 & 34.6 & 26.4 & 21.7 & 20.5 & 19.6 & 32.7\\
        % Large$^{\ddagger}$ & 61.0 & 60.1 & 54.2 & 35.8 & 27.5 & 22.9 & 22.4 & 21.6 & 38.2 \\

        % \flickr small &  \\
        % Hybrid Large & 67.8  & 39.7 \\
        % Hybrid Large & 77.7 & 77.6 & 59.3 & 43.7 & 39.1 & 38.2 & 27.5 & 16.2 & 47.4 \\
        % Cascaded Large  & 27.5 &	22.4 &	35.8 &	61.0 &	21.6 &	54.2 &	60.1 &	22.9 \\
        % Hybrid Large & 66.7 &	16.1 &	58.1 &	18.3 &	55.3 &	35.2 &	67.8 &	54.7 \\
        % Hybrid-WS Large & 77.7 &	16.2 &	43.7 &	27.5 &	38.2 &	39.1 &	77.6 &	59.3 \\
        \bottomrule
    \end{tabular}
    \vspace*{-12pt}
\end{table}

Due to the unique design of cascaded \speechclip, we investigate what and how well the speech encoder extracts keywords.
For each encoded and normalized CLS token $\boldsymbol{z}_k$, keywords are retrieved by finding subwords with the highest cosine similarities between $\boldsymbol{z}_k$ and the corresponding subword embeddings.
Notice that previous works~\cite{kamper2018semantic,pasad2019contributions} are also capable of retrieving semantically related keywords from speech. Nonetheless, they required pretrained image tagger and the size of keywords set is very limited. For \speechclip, we can apply the same method to other pretrained langange models' vocabulary, technically.
Also, our setting is quite different from ~\cite{olaleye2022keyword}, where the 8 keywords are discovered from speech utterance without any text query in our work. Namely, \speechclip can automatically summarize the speech by selecting 8 keywords.
We offer quantitative and qualitative analyses in the following paragraphs.

We inspect how well keywords are retrieved from speech signals for the quantitative analysis.
The evaluation metric is hit rate, which is the percentage of successful top-1 keyword retrieval of any word in the caption averaged over all testing samples.
In Table~\ref{tab:kw_hit}, some CLS tokens frequently retrieve words in the ground truth captions, showing that the cascaded architecture can directly capture words from speech.
Moreover, the first keyword's hit rate for models trained on \flickr is relatively high compared to other keywords.
Probably because the first word in a sentence has a higher chance to be ``a'', which is also the top-1 commonly retrieved subword from the first keyword in \flickr.
Another finding is that the Large model obtains a higher averaged keyword hit rate than the Base model on \flickr, which is consistent with the trend in Table~\ref{tab:recall}.
Hence, retrieving correct keywords is related to retrieving between speech and image samples.
Although some CLS tokens obtain reasonable hit rates, one might question whether the retrieved words are meaningful instead of stopwords.
Hence, we next analyze the results qualitatively to address this concern.

For the qualitative analysis, we offer two samples from the \coco testing set in Fig.~\ref{fig:kw_ret}, showing their attention maps in the transformer encoder and retrieved words for each CLS token.
In the first example, although only a few retrieved keywords are in the ground truth caption, some semantically related words are found.
For instance, attention maps of keywords 1, 2, and 6 focus on segments uttering ``tie'' and ``suit.''
Meanwhile, they retrieve words related to clothes and appearance, e.g., ``dapper'', ``tuxedo'', and ``scarf.''
A similar trend can be found in the second sample, showing that the cascaded objective makes the speech encoder captures semantic information.
Moreover, looking at both examples, each keyword seems to have a particular purpose, e.g., the 8$\text{th}$ keyword tends to retrieve specific nouns from utterances while the 7$\text{th}$ retrieves prepositions.
This observation leads us to investigate the properties of each keyword.

% \begin{table}[t]
%     \caption{
%         Common retrieved subowords for each keyword on \coco test set.
%     }
%     \vspace{2pt}
%     \label{tab:kw_common_ret}
%     \centering
%     \small
%     % \footnotesize
%     \begin{tabular}{@{~}l@{~~}l}
%         \toprule
%         \# & Common Retrieved Subwords \\
%         \midrule
%     1 & a, pizza, the, giraffe, bathroom, skateboard, living, gira \\
%     2 & cat, a, room, sheep, frisbee, skis, bird, skateboard \\
%     3 & bathroom, skateboard, room, horse, elephant, motorcycle, kitchen, clock \\
%     4 & a, of, in, man, woman, dog, train, with \\
%     5 & a, tennis, with, eating, and, playing, the, flying \\
%     6 & street, bathroom, kitchen, train, beach, bed, bus, grass \\
%     7 & in, of, to, from, for, a, on, at \\
%     8 & train, sign, cake, clock, is, bus, truck, car \\
%         \bottomrule
%     \end{tabular}
%     % \vspace*{-8pt}
% \end{table}

\begin{table}[t]
    \caption{
        Top 10 successfully retrieved subwords for each keyword on \coco test set using the cascaded Large model.
        The subwords are sorted in decreasing occurrence.}
    % \vspace{2pt}
    \label{tab:kw_common_ret}
    \centering
    % \small
    \footnotesize
    % \begin{tabular}{@{~}c@{~}c@{~}c@{~}c@{~}c@{~}c@{~}c@{~}c@{~}c@{~}}
    \begin{tabular}{@{~}l@{~}l@{~}l@{~}l@{~}l@{~}l@{~}l@{~}l@{~}l@{~}}
        \toprule
         kw1 & kw2 & kw3 & kw4 & kw5 & kw6 & kw7 & kw8 \\
        \midrule
        a & cat & bathroom & a & a & street & in & train \\
        pizza & a & skateboard & of & tennis & bathroom & of & sign \\
        the & room & room & in & with & kitchen & to & cake \\
        giraffe & sheep & horse & man & eating & train & from & clock \\
        bathroom & frisbee & elephant & woman & and & beach & for & is \\
        skateboard & skis & motorcycle & dog & playing & bed & a & bus \\
        living & bird & kitchen & train & the & bus & on & truck \\
        gira & skateboard & clock & with & flying & grass & at & car \\
        sheep & surf & tower & is & sitting & road & the & of \\
        an & kite & bear & to & walking & room & -- & signs \\
        \bottomrule
    \end{tabular}
    \vspace*{-12pt}
\end{table}

In Table~\ref{tab:kw_common_ret}, we list the top 10 successfully and frequently retrieved subwords for each keyword in \coco.
Generally, commonly retrieved subwords are either stopwords like ``a'' and ``of'' or objects like ``skateboard'' and ``street.''
In the first case, the phenomenon might be caused by the supervision from the CLIP text encoder because stopwords contain little information about speech signals but are sometimes crucial for maintaining the syntactic structures.
Moreover, we find the frequently retrieved words for objects sometimes appear in \coco's captions but not very frequently.
Hence, these words might be easier to be detected in speech, and the corresponding objects are more concrete to be found in images.

Additionally, we find that some keywords predict specific subword categories successfully.
For instance, keyword 7 tends to output prepositions and articles, while keyword 5 mostly retrieves action words.
As for the rest of the keywords, nouns are mostly retrieved.
Particularly, for keyword 2, ``frisbee'', ``skis'', ``skateboard'', and ``surf'' are all related to outdoor activities.
As for keyword 8, ``train'', ``sign'', ``bus'', ``truck'', ``car'', and ``signs'' are all related to traffic.
This section demonstrates the cascaded \speechclip for retrieving semantically related keywords from speech signals.

\subsection{Layer Importance in \speechclip Speech Encoder}
\label{subsubsec:speechCLIP_weights}

\begin{figure}
    \centering
    \includegraphics[width=\linewidth,trim={0.4cm 0.2cm 1.0cm 0.3cm},clip]{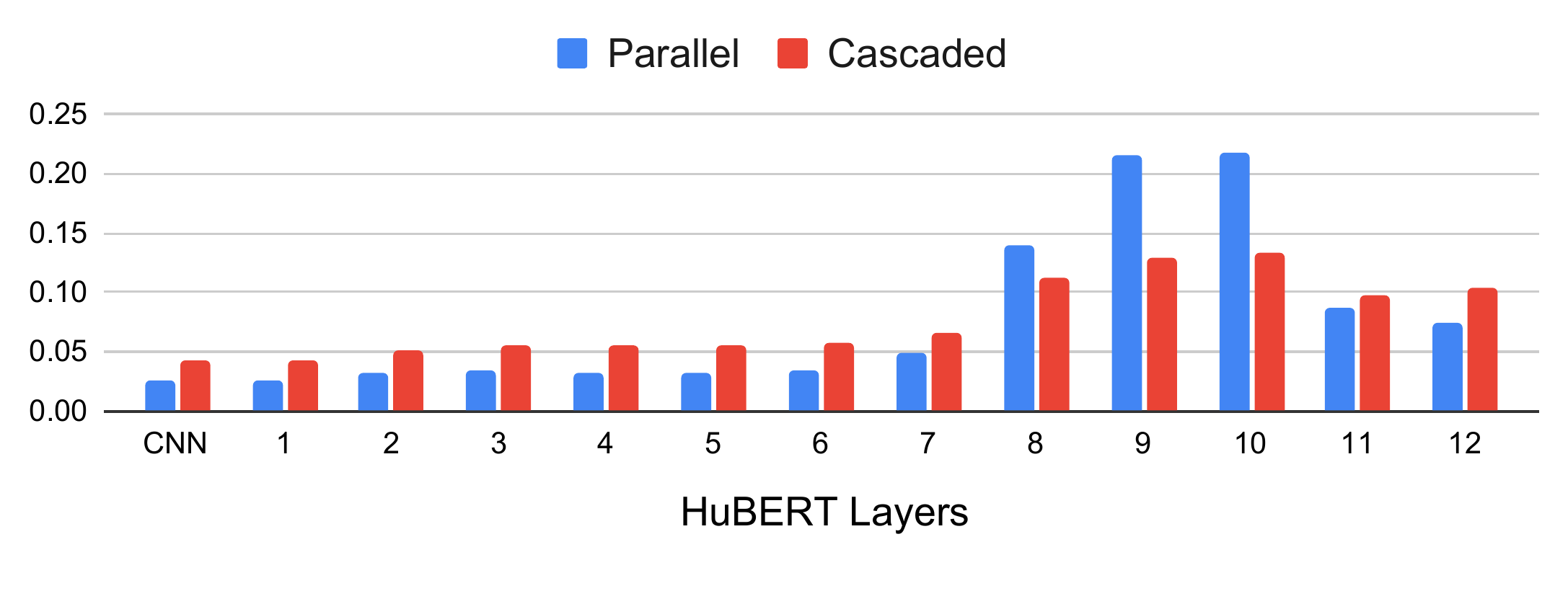}
    \vspace{-28pt}
    \caption{
        Normalized weights for layer summarization of \hubert  in parallel and cascaded \speechclip .
        \texttt{CNN} denotes the \hubert CNN feature extractor.
    }
    \label{fig:speechCLIP_weight}
    \vspace*{-8pt}
\end{figure}

% In previous sections, we have shown \speechclip's abilities for zero-shot speech-text retrieval and keywords extractions, which is also worth knowing how \speechclip leverages \hubert's hidden representations.
In this section, we show which \hubert hidden layers are crucial for \speechclip to perform well in various tasks discussed earlier.
Hence, we visualize the learned weights in the weighted sum mechanism mentioned in Sec.~\ref{subsec:prelim} in Fig.~\ref{fig:speechCLIP_weight}.
Both parallel and cascaded \speechclip utilize the roughly the 8$^{\text{th}}$ to the 10$^{\text{th}}$ layers in \hubert, inferring that \hubert's top layers capture rich content and semantic information.
This result is consistent with prior works investigating the importance of different hidden layers in speech SSL models~\cite{baevski2021wav2vec-u,pasad2021layer,chang2022distilhubert}, i.e., the top hidden layers contain word meaning and content information.
However, the cascaded model's weights distribute more evenly over the layers than parallel \speechclip, showing that the model architecture design affects the utilization of \hubert's layers.
% Overall, because of the retrieval objective in \speechclip, the two variants leverage \hubert's hidden representations with rich semantic
% As for the lower layers, they might focus on acoustics information, so in this case, do not get substantial weights.

% Next, we inspect the \speechclip objectives' similarity to other speech processing problems.
% Because we follow SUPERB~\cite{yang2021superb} for weighted sum and freeze speech SSL models' hidden representations, we can compare the learned weights for solving different tasks in the same \hubert model.
% The learned weights are normalized and can be seen as a probability distribution over 13 different categories.
% Then, we calculate the Jensen–Shannon~(JS) divergence between each the learned weights in different tasks.
% Lower values indicate higher similarities, which also imply that the tasks utilize \hubert's hidden representations in a similar manner.

% perhaps for the ICASSP paper
% \input{figure/layer_weights_SUPERB}

% From Fig.~\ref{fig:superb_layer_weights}, we see that Keyword Spotting (KS) has the closest weight distribution compared to our \speechclip, especially for cascaded model.
% This is quite reasonable since cascaded model are trained to predict keywords, which resembles keyword spotting to some extent.
% In addition, of all the SUPERB tasks, Speaker Identification (SID) has the most different distribution compared to all the other settings.

\subsection{Ablation Studies}
\label{subsec:ablation}
\begin{table}[t]
    \centering
    \caption{
        Recall scores on \flickr for ablation studies.
    }
    \label{tab:ablation}
    \small
    \vspace{-5pt}
    \begin{tabular}{@{~~}l@{~~}c@{~~}c@{~~}c@{~}c@{~}c@{~~}c@{~~}c@{~~}} 
        \toprule
        & \multicolumn{3}{c}{Speech $\rightarrow$ Image} & & \multicolumn{3}{c}{Image $\rightarrow$ Speech} \\ 
        \cmidrule{2-4}
        \cmidrule{6-8}
        Method & R@1 & R@5 & R@10 & & R@1 & R@5 & R@10 \\
        \midrule
        & \multicolumn{7}{c}{Batch Normalization} \\
        \cmidrule{2-8}
        Cascaded (w/ BN) & 8.2  & 25.7 & 37.2 & & 14.1 & 34.5 & 49.2 \\
        Cascaded (w/o BN) & 1.1 & 4.7 & 8.4 & & 1.4 & 5.7 & 9.4 \\
        \midrule
        & \multicolumn{7}{c}{Keyword Num} \\
        \cmidrule{2-8}
        Cascaded ($K$ $=$ 8) & 8.2  & 25.7 & 37.2 & & 14.1 & 34.5 & 49.2 \\
        Cascaded ($K$ $=$ 4) & 3.5 & 13.2  & 21.1 & & 5.2 & 17.5 & 27.4 \\
        Cascaded ($K$ $=$ 2) & 2.1 & 8.4   & 14.4 & & 2.7 & 10.6 & 17.6 \\

        % \hl{Hybrid$_{\text{C}}$ Large} ($K=8$) & 15.2 & 39.8 & 54.6 & & 19.2 & 50.2 &	63.5 \\
        % \hl{Hybrid$_{\text{C}}$ Large} ($K=4$) & 2.0 &	8.2 &	14.2 & & 2.9 & 9.8 &	16.5 \\
        % \hl{Hybrid$_{\text{C}}$ Large} ($K=2$) & 2.1 &	8.7 &	14.9 & & 3.0 & 10.8 &	17.7\\
        
        % \midrule
        % & \multicolumn{7}{c}{Learnable Temperature} \\
        % \cmidrule{2-8}
        % Parallel Large & 35.6 & 64.3 & 75.9 & & 47.9 & 77.6 & 86.9 \\
        % \hl{Parallel Large (Fixed)} & 25.6 & 53.2 & 66.5 & & 46.8 & 75.8 &	85.3 \\
        
        \bottomrule
    \end{tabular}
    \vspace*{-10pt}
\end{table}

\subsubsection{Batch Normalization in Cascaded \speechclip}
\label{subsubsec:batchnorm}

Here, we demonstrate the importance of batch normalization in the cascaded \speechclip.
We compare cascaded \speechclip with its variant without using batch normalization, as shown in the first two rows of Table~\ref{tab:ablation}.
Removing batch normalization degrades retrieval performance significantly, showing the significance of mean and variance matching described in Sec.~\ref{subsec:cascaded}.

\subsubsection{Number of Keywords in Cascaded \speechclip}
\label{subsubsec:kwnum}

This section discusses the impact of the number of keywords in cascaded \speechclip.
We report retrieval results on \flickr using different amounts of keywords in Table~\ref{tab:ablation}.
Results show that reducing keywords degrades retrieval performance, indicating that using fewer keywords is incapable of passing information from the speech encoder to the CLIP text encoder.
Furthermore, the number of subword tokens in a \flickr utterance is 11.3 $\pm$ 4.1, and some tokens carry less information like stopwords.
Therefore, we suggest 8 is a reasonable number for $K$ to obtain good performance with cascaded \speechclip.
Although dynamically assigning $K$ for utterances of different lengths is more appropriate, we leave this approach for future investigation.

% Also, notice that the length of subword tokens on SpokenCOCO is $10.7\pm2.5$, and we know that not all of the subwords in the captions are crucial for semantic and content information.

% \subsubsection{Temperature Tuning in Contrastive Learning}
% \label{subsubsec:learnTemp}

% This section shows the importance of learnable temperature in contrastive loss.
% In the last section of Table~\ref{tab:ablation}, removing the learnable temperature in the contrastive objective degrades recall scores more than 10\%.
% Hence, the experiments show 
\vspace{-5pt}

\section{Conclusion}
\label{sec:conclusion}

This paper introduces \speechclip, a novel framework integrating CLIP into visually grounded speech models.
We demonstrate significant improvements in image-speech retrieval with CLIP's supervision.
Moreover, the proposed methods can perform zero-shot speech-text retrieval and capture semantically related keywords in speech signals.
Results indicate that bridging speech and text domains with CLIP's supervision is possible and promising.
Overall, \speechclip opens a new research direction of indirectly supervising speech models with text via other modalities.
We suggest some topics in \speechclip are worth investigating in the future, including integrating parallel and cascaded in the same model and cascaded structure with variable length prediction aiming for unsupervised ASR.
% transcribing speech into text sentences.
Furthermore, extending \speechclip to a multilingual model is possible using spoken captions from other languages or Multilingual-CLIP models.
Finally, we wish to inspect how CLIP can enhance speech SSL models' performance on downstream problems like speech recognition and intent classification.

% Moreover, if we have spoken captions for other non-English languages, we could apply \speechclip to learn English text translation from of non-English speech.
% We suggest some topics in \speechclip should be investigated in the future, including extending to other languages, \david{We could mention applying cascaded speechCLIP to learn the English text translations of non-English speech in the case that we have spoken captions for other languages} combining parallel and cascaded objectives, and proposing a cascaded architecture with variable length prediction aiming for transcribing speech into text.
\vspace{-5pt}
\section{ACKNOWLEDGMENTS}
\label{sec:ack}

% Also, this work is done during the 2022 Eighth Frederick Jelinek Memorial Summer Workshop.
% This work was supported by JSALT 2022 at JHU, with gift-funds from Amazon, Microsoft, and Google.
% Also, we thank Taiwan Web Service (TWS) for providing computational resources.
This work was supported by JSALT 2022 at JHU, with gift-funds from Amazon, Microsoft, and Google.
Also, we thank Taiwan Web Service (TWS) and National Center for High-performance Computing (NCHC) of National Applied Research Laboratories (NARLabs) for providing computational resources.

\bibliographystyle{IEEEbib}
\bibliography{refs}

\begin{thebibliography}{10}

\bibitem{mohamed2022sslreview}
Abdelrahman Mohamed, Hung-yi Lee, Lasse Borgholt, Jakob~D Havtorn, Joakim Edin,
  Christian Igel, Katrin Kirchhoff, Shang-Wen Li, Karen Livescu, Lars
  Maal{\o}e, et~al.,
\newblock ``Self-supervised speech representation learning: A review,''
\newblock {\em arxiv:2205.10643}, 2022.

\bibitem{liu2020mockingjay}
Andy~T Liu, Shu-wen Yang, Po-Han Chi, Po-chun Hsu, and Hung-yi Lee,
\newblock ``Mockingjay: Unsupervised speech representation learning with deep
  bidirectional transformer encoders,''
\newblock in {\em ICASSP}, 2020.

\bibitem{liu2021tera}
Andy~T Liu, Shang-Wen Li, and Hung-yi Lee,
\newblock ``{TERA}: Self-supervised learning of transformer encoder
  representation for speech,''
\newblock {\em TASLP}, vol. 29, 2021.

\bibitem{chung2019apc}
Yu-An Chung, Wei-Ning Hsu, Hao Tang, and James Glass,
\newblock ``An unsupervised autoregressive model for speech representation
  learning,''
\newblock in {\em Interspeech}, 2019.

\bibitem{chung2020vq-apc}
Yu-An Chung, Hao Tang, and James Glass,
\newblock ``Vector-quantized autoregressive predictive coding,''
\newblock in {\em Interspeech}, 2020.

\bibitem{liu2021npc}
Alexander~H Liu, Yu-An Chung, and James Glass,
\newblock ``Non-autoregressive predictive coding for learning speech
  representations from local dependencies,''
\newblock in {\em Interspeech}, 2021.

\bibitem{oord2018cpc}
Aaron van~den Oord, Yazhe Li, and Oriol Vinyals,
\newblock ``Representation learning with contrastive predictive coding,''
\newblock {\em arxiv:1807.03748}, 2018.

\bibitem{schneider2019wav2vec}
Steffen Schneider, Alexei Baevski, Ronan Collobert, and Michael Auli,
\newblock ``wav2vec: Unsupervised pre-training for speech recognition,''
\newblock in {\em Interspeech}, 2019.

\bibitem{baevski2020vq-wav2vec}
Alexei Baevski, Steffen Schneider, and Michael Auli,
\newblock ``vq-wav2vec: Self-supervised learning of discrete speech
  representations,''
\newblock in {\em ICLR}, 2020.

\bibitem{baevski2020wav2vec2}
Alexei Baevski, Yuhao Zhou, Abdelrahman Mohamed, and Michael Auli,
\newblock ``wav2vec 2.0: A framework for self-supervised learning of speech
  representations,''
\newblock in {\em NeurIPS}, 2020.

\bibitem{chung2021w2v}
Yu-An Chung, Yu~Zhang, Wei Han, Chung-Cheng Chiu, James Qin, Ruoming Pang, and
  Yonghui Wu,
\newblock ``W2v-bert: Combining contrastive learning and masked language
  modeling for self-supervised speech pre-training,''
\newblock in {\em ASRU}, 2021.

\bibitem{hsu2021hubert}
Wei-Ning Hsu, Benjamin Bolte, Yao-Hung~Hubert Tsai, Kushal Lakhotia, Ruslan
  Salakhutdinov, and Abdelrahman Mohamed,
\newblock ``{HuBERT}: Self-supervised speech representation learning by masked
  prediction of hidden units,''
\newblock {\em arxiv:2106.07447}, 2021.

\bibitem{chen2021wavlm}
Sanyuan Chen, Chengyi Wang, Zhengyang Chen, Yu~Wu, Shujie Liu, Zhuo Chen, Jinyu
  Li, Naoyuki Kanda, Takuya Yoshioka, Xiong Xiao, et~al.,
\newblock ``Wavlm: Large-scale self-supervised pre-training for full stack
  speech processing,''
\newblock {\em arxiv:2110.13900}, 2021.

\bibitem{chiu2022bestrq}
Chung-Cheng Chiu, James Qin, Yu~Zhang, Jiahui Yu, and Yonghui Wu,
\newblock ``Self-supervised learning with random-projection quantizer for
  speech recognition,''
\newblock {\em ICML}, 2022.

\bibitem{ravanelli2020paseplus}
Mirco Ravanelli, Jianyuan Zhong, Santiago Pascual, Pawel Swietojanski, Joao
  Monteiro, Jan Trmal, and Yoshua Bengio,
\newblock ``Multi-task self-supervised learning for robust speech
  recognition,''
\newblock in {\em ICASSP}, 2020.

\bibitem{chang2022distilhubert}
Heng-Jui Chang, Shu-wen Yang, and Hung-yi Lee,
\newblock ``Distilhubert: Speech representation learning by layer-wise
  distillation of hidden-unit bert,''
\newblock in {\em ICASSP}, 2022.

\bibitem{baevski2022data2vec}
Alexei Baevski, Wei-Ning Hsu, Qiantong Xu, Arun Babu, Jiatao Gu, and Michael
  Auli,
\newblock ``Data2vec: A general framework for self-supervised learning in
  speech, vision and language,''
\newblock {\em arxiv:2202.03555}, 2022.

\bibitem{wang2022lighthubert}
Rui Wang, Qibing Bai, Junyi Ao, Long Zhou, Zhixiang Xiong, Zhihua Wei,
  Yu~Zhang, Tom Ko, and Haizhou Li,
\newblock ``Lighthubert: Lightweight and configurable speech representation
  learning with once-for-all hidden-unit bert,''
\newblock {\em arxiv:2203.15610}, 2022.

\bibitem{yang2021superb}
Shu-wen Yang, Po-Han Chi, Yung-Sung Chuang, Cheng-I~Jeff Lai, Kushal Lakhotia,
  Yist~Y Lin, Andy~T Liu, Jiatong Shi, Xuankai Chang, Guan-Ting Lin, et~al.,
\newblock ``{SUPERB}: Speech processing universal performance benchmark,''
\newblock in {\em Interspeech}, 2021.

\bibitem{evain2021lebenchmark}
Sol\`ene~Evain \textit{et al.},
\newblock ``{LeBenchmark}: A reproducible framework for assessing
  self-supervised representation learning from speech,''
\newblock in {\em Interspeech}, 2021.

\bibitem{tsai2022superbsg}
Hsiang-Sheng~Tsai \textit{et al.},
\newblock ``{SUPERB}-{SG}: Enhanced speech processing universal {PER}formance
  benchmark for semantic and generative capabilities,''
\newblock in {\em ACL}, 2022.

\bibitem{harwath2018jointly}
David Harwath, Adria Recasens, D{\'\i}dac Sur{\'\i}s, Galen Chuang, Antonio
  Torralba, and James Glass,
\newblock ``Jointly discovering visual objects and spoken words from raw
  sensory input,''
\newblock in {\em ECCV}, 2018.

\bibitem{radford2021clip}
Alec~Radford \textit{et al.},
\newblock ``Learning transferable visual models from natural language
  supervision,''
\newblock in {\em ICML}, 2021.

\bibitem{chrupala2022visually}
Grzegorz Chrupa{\l}a,
\newblock ``Visually grounded models of spoken language: A survey of datasets,
  architectures and evaluation techniques,''
\newblock {\em JAIR}, vol. 73, 2022.

\bibitem{hsu2019transfer}
Wei-Ning Hsu, David Harwath, and James Glass,
\newblock ``Transfer learning from audio-visual grounding to speech
  recognition,''
\newblock {\em Interspeech}, 2019.

\bibitem{harwath2015deep}
David Harwath and James Glass,
\newblock ``Deep multimodal semantic embeddings for speech and images,''
\newblock in {\em ASRU}, 2015.

\bibitem{hsu2020text}
Wei-Ning Hsu, David Harwath, Christopher Song, and James Glass,
\newblock ``Text-free image-to-speech synthesis using learned segmental
  units,''
\newblock {\em arxiv:2012.15454}, 2020.

\bibitem{wang2021align}
Liming Wang, Xinsheng Wang, Mark Hasegawa-Johnson, Odette Scharenborg, and
  Najim Dehak,
\newblock ``Align or attend? toward more efficient and accurate spoken word
  discovery using speech-to-image retrieval,''
\newblock in {\em ICASSP}, 2021.

\bibitem{khorrami2021evaluation}
Khazar Khorrami and Okko R{\"a}s{\"a}nen,
\newblock ``Evaluation of audio-visual alignments in visually grounded speech
  models,''
\newblock {\em Interspeech}, 2021.

\bibitem{harwath2018vision}
David Harwath, Galen Chuang, and James Glass,
\newblock ``Vision as an interlingua: Learning multilingual semantic embeddings
  of untranscribed speech,''
\newblock in {\em ICASSP}, 2018.

\bibitem{kamper2018visually}
Herman Kamper and Michael Roth,
\newblock ``Visually grounded cross-lingual keyword spotting in speech,''
\newblock {\em SLTU}, 2018.

\bibitem{havard2020catplayinginthesnow}
William~N Havard, Jean-Pierre Chevrot, and Laurent Besacier,
\newblock ``Catplayinginthesnow: Impact of prior segmentation on a model of
  visually grounded speech,''
\newblock {\em CoNLL}, 2020.

\bibitem{ohishi2020trilingual}
Yasunori Ohishi, Akisato Kimura, Takahito Kawanishi, Kunio Kashino, David
  Harwath, and James Glass,
\newblock ``Trilingual semantic embeddings of visually grounded speech with
  self-attention mechanisms,''
\newblock in {\em ICASSP}, 2020.

\bibitem{ilharco2019large}
Gabriel Ilharco, Yuan Zhang, and Jason Baldridge,
\newblock ``Large-scale representation learning from visually grounded
  untranscribed speech,''
\newblock {\em CoNLL}, 2019.

\bibitem{sanabria2021milan}
Ramon Sanabria, Austin Waters, and Jason Baldridge,
\newblock ``Talk, don't write: A study of direct speech-based image
  retrieval,''
\newblock {\em Interspeech}, 2021.

\bibitem{peng2022fastvgs}
Puyuan Peng and David Harwath,
\newblock ``Fast-slow transformer for visually grounding speech,''
\newblock in {\em ICASSP}, 2022.

\bibitem{peng2022fastvgsplus}
Puyuan Peng and David Harwath,
\newblock ``Self-supervised representation learning for speech using visual
  grounding and masked language modeling,''
\newblock {\em arxiv:2202.03543}, 2022.

\bibitem{peng2022vghubert}
Puyuan Peng and David Harwath,
\newblock ``Word discovery in visually grounded, self-supervised speech
  models,''
\newblock {\em Interspeech}, 2022.

\bibitem{pasad2021layer}
Ankita Pasad, Ju-Chieh Chou, and Karen Livescu,
\newblock ``Layer-wise analysis of a self-supervised speech representation
  model,''
\newblock {\em arxiv:2107.04734}, 2021.

\bibitem{baevski2021wav2vec-u}
Alexei Baevski, Wei-Ning Hsu, Alexis Conneau, and Michael Auli,
\newblock ``Unsupervised speech recognition,''
\newblock {\em NeurIPS}, 2021.

\bibitem{liu2022wav2vec-u2}
Alexander~H Liu, Wei-Ning Hsu, Michael Auli, and Alexei Baevski,
\newblock ``Towards end-to-end unsupervised speech recognition,''
\newblock {\em arxiv:2204.02492}, 2022.

\bibitem{wu2022wav2clip}
Ho-Hsiang Wu, Prem Seetharaman, Kundan Kumar, and Juan~Pablo Bello,
\newblock ``Wav2clip: Learning robust audio representations from clip,''
\newblock in {\em ICASSP}, 2022.

\bibitem{guzhov2022audioclip}
Andrey Guzhov, Federico Raue, J{\"o}rn Hees, and Andreas Dengel,
\newblock ``Audioclip: Extending clip to image, text and audio,''
\newblock in {\em ICASSP}, 2022.

\bibitem{vaswani2017attention}
Ashish Vaswani, Noam Shazeer, Niki Parmar, Jakob Uszkoreit, Llion Jones,
  Aidan~N Gomez, {\L}ukasz Kaiser, and Illia Polosukhin,
\newblock ``Attention is all you need,''
\newblock in {\em NeurIPS}, 2017.

\bibitem{bengio2013estimating}
Yoshua Bengio, Nicholas L{\'e}onard, and Aaron Courville,
\newblock ``Estimating or propagating gradients through stochastic neurons for
  conditional computation,''
\newblock {\em arxiv:1308.3432}, 2013.

\bibitem{karpathy2015deep}
Andrej Karpathy and Li~Fei-Fei,
\newblock ``Deep visual-semantic alignments for generating image
  descriptions,''
\newblock in {\em CVPR}, 2015.

\bibitem{duquenne2021multimodal}
Paul-Ambroise Duquenne, Hongyu Gong, and Holger Schwenk,
\newblock ``Multimodal and multilingual embeddings for large-scale speech
  mining,''
\newblock in {\em NeurIPS}, 2021.

\bibitem{khurana2022samu}
Sameer Khurana, Antoine Laurent, and James Glass,
\newblock ``{SAMU-XLSR}: Semantically-aligned multimodal utterance-level
  cross-lingual speech representation,''
\newblock {\em arxiv preprint}, 2022.

\bibitem{kamper2018semantic}
Herman Kamper, Gregory Shakhnarovich, and Karen Livescu,
\newblock ``Semantic speech retrieval with a visually grounded model of
  untranscribed speech,''
\newblock in {\em CVPR}, 2018, pp. 2514--2517.

\bibitem{pasad2019contributions}
Ankita Pasad, Bowen Shi, Herman Kamper, and Karen Livescu,
\newblock ``On the contributions of visual and textual supervision in
  low-resource semantic speech retrieval,''
\newblock {\em arXiv preprint}, 2019.

\bibitem{olaleye2022keyword}
Kayode Olaleye, Dan Oneata, and Herman Kamper,
\newblock ``Keyword localisation in untranscribed speech using visually
  grounded speech models,''
\newblock {\em JSTSP}, 2022.

\end{thebibliography}

\end{document}